\def\year{2019}\relax
\documentclass[letterpaper]{article} 
\usepackage{aaai19}  
\usepackage{times}  
\usepackage{helvet}  
\usepackage{courier}  
\usepackage{url}  
\usepackage{graphicx}  
\usepackage{amsmath}
\usepackage{subcaption}
\begin{document}
%
\title{ A Compositional Textual Model for Recognition of Imperfect Word Images}
 \author{
  Wei Tang\thanks{Equal contribution}, John Corring\footnotemark[1], Ying Wu, Gang Hua
}

\maketitle
\begin{abstract}
Printed text recognition is an important problem for industrial OCR systems. Printed text is constructed in a standard procedural fashion in most settings. We develop a mathematical model for this process that can be applied to the backward inference problem of text recognition from an image. Through ablation experiments we show that this model is realistic and that a multi-task objective setting can help to stabilize estimation of its free parameters, enabling use of conventional deep learning methods. Furthermore, by directly modeling the geometric perturbations of text synthesis we show that our model can help recover missing characters from incomplete text regions, the bane of multicomponent OCR systems, enabling recognition even when the detection returns incomplete information.
\end{abstract}

\section{Introduction}
Automated visual text recognition is a fundamental problem of computer vision, with a history as old as the subject itself.~\cite{bledsoe1959pattern} Core to the problem are the degrees of variation within the text construction process itself: what kind of font is used, with what kerning (spacing), on what surface, etc.? These considerations have been explored in prior works on document modeling~\cite{DBLP:conf/icip/KopecC94} and document cleaning ~\cite{DBLP:journals/pami/DaiL14}, but we have not seen them explored specifically for the problem of word image recognition.

In this work we attempt to model inverse inference for the word image construction process using modern day techniques, primarily deep neural networks. We outline a model for word image formation and then propose modules to invert each component in the compositional model. As far as we are aware, we are the first to employ compositional modeling for the new regime of CTC-based word recognition algorithms. The task we focus on is the backward inference task of recognizing the text in the image, rather than generative modeling. 

The most recent wave of successful techniques~\cite{DBLP:journals/corr/ShiBY15,DBLP:journals/corr/ShiWLYB16,DBLP:conf/aaai/HeH0LT16,DBLP:conf/aaai/LiuCW18} stem from formulation of the inference loss by means of connectionist temporal classification (CTC)~\cite{DBLP:conf/icml/GravesFGS06}. This approach eschews explicit segmentation of the text images into character regions and solves the many to many problem by decoding the predictions with dynamic programming. This formulation dictates the latter part of almost all text recognition algorithms. While some alternatives have arisen recently, such as edit probability based decoding \cite{DBLP:journals/corr/abs-1805-03384}, we focus on CTC-based decoding in this work. In principle, most of our work can be adapted to most segmentation free decoding strategies.

Before the advent of the CTC-based approaches, character segmentation saw several promising avenues of research~\cite{DBLP:journals/pami/LeeLP96,DBLP:conf/icdar/PhanSST11}. As a consequence of adapting the segmentation-free framework of CTC, few authors have explored the variation of kerning or character segmentation within word images.

Modeling geometric nuisances in classification and recognition also has a long history in Computer Vision. Keypoint-based registration methods have been employed in the past~\cite{DBLP:conf/iccv/PhanSTT13}. However, it has only recently been rediscovered in deep learning frameworks~\cite{DBLP:conf/nips/JaderbergSZK15,DBLP:conf/eccv/ChenHW016,DBLP:conf/cvpr/LinL17}. In the current trend of text recognition, few works have explored canonical alignment of word images~\cite{DBLP:journals/corr/ShiWLYB16,DBLP:conf/bmvc/LiuCWSH16}. Both of these works employ 2-stage training regimes, and ~\cite{DBLP:journals/corr/ShiWLYB16} does not do pixel-based alignment. Meanwhile, character-based alignment suffers from the segmentation problem~\cite{DBLP:conf/aaai/LiuCW18}, which the CTC approach is designed to avoid. This makes this solution for character alignment intuitively unappealing, since failure to segment correctly will limit the registration and thus the recognition. While the works above model nuisances {\it within} the region attended by the detector, we are not aware of works that model both the geometric noise of the text {\bf and} the noise of the detector, as we do in this work. As a result, our work is well suited for 2-stage end-to-end systems.

This work is not focused on development of a full end-to-end system, but rather a second stage that works well independent of the first stage. Thus, we show that our approach to text recognition leads to a more robust 2-stage pipeline and thus we compare with others as well as internally in this problem setting. 

\subsection{Contributions}

We are the first to formulate a model for text recognition that enables end-to-end alignment between the recognizer and a detector trained independently. We are also the first to show that a template reconstruction loss can be used to stabilize the training of spatial transformer networks for text recognition, enabling a single joint training. 

It is long been posited that compositional modeling provides the clearest route towards disentangling image representation across their factors of variation.~\cite{DBLP:journals/ftcgv/ZhuM06,zhu2011recursive} To the best of our knowledge, we have provided the first complete and concise compositional model for word image formation applied to this problem. Our work can be used for end-to-end groundtruth that has different levels of precision (axis aligned, oriented bounding box, or quadrilateral) all in the same data set. Finally, we provide experiments that show that each module serves an essential role in the final inference problem. The result is an algorithm that is competitive with state of the art recognizers and can be plugged into detectors trained independently and yields top-notch end-to-end performance.

\section{Forward Compositional Textual Model}

Our approach to rectification and recognition leverages an explicit decomposition of the text formation process into 5 distinct steps:
\begin{enumerate}
\item Transcription and skeletonization,
\item Kerning,
\item Typesetting (font),
\item Lighting, shading, and background distortion (appearance),
\item Geometric distortion.
\end{enumerate}
We explicitly model each step in this process and show that our model leads to improved performance from the CRNN baseline with ablation studies.

Bear in mind that the primary task of our work is to invert the text rendering process. We describe each component in its forward form, and finally we formalize our task in terms of the inversion of these functions.

\subsection{Transcription and skeletonization}

Together, transcription, skeletonization, font, and kerning are the intrinsic parameters of the rendered text while geometric distortion and appearance are extrinsic. In this subsection we discuss transcription and skeletonization.

In our setting, {\it Transcription} is the process of converting an atomic word symbol into a sequence of character symbols that will be rendered. Thus, transcription converts a word symbol from a dictionary, $w \in \mathcal{D}$, to a sequence of characters from an alphabet, $s_W \in \Sigma^\ast$, by a function $T: \mathcal{D} \rightarrow \Sigma^\ast$. 

{\it Skeletonization} is the process of converting a sequence of character symbols into a canonical word image.  We refer to the step of rendering a canonical word image as skeletonization because we choose to use character skeletons to represent visual character symbols. Thus, skeletonization converts a sequence of characters from an alphabet, $s_W \in \Sigma^\ast$, so an image, $s: \Sigma^\ast \rightarrow B(I)$ where $B(I)$ is the set of bounded functions on the spatial image domain $I$. We replace $s \circ T$ with $s$ below, assuming that the input will be in the form of a character sequence.

\subsection{Kerning}

\begin{figure}
\centering{
\includegraphics[width=.5\linewidth]{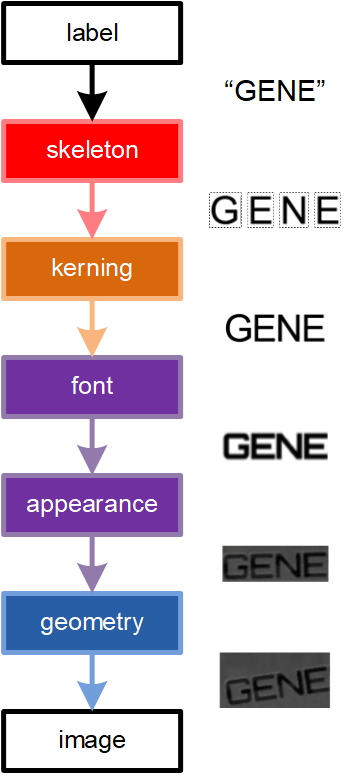}
}
\caption{A schematic visualization of the text generation process we model our inference network on. Transcription is not pictured here, because it is the generation of the word sequence itself. }
\label{fig:completemodel}
\end{figure}

The function $s$ does not directly model {\it kerning}, the spacing between characters, which can be independent of font. We model the kerning operation as a function of the image produced by $s(w)$. It converts an image with one type of spacing to an image with another spacing, which may be nonuniform between characters. The kerning function can be different for identical word inputs, thus we model this variability with free parameters, $\theta \in \Theta_k$. Thus $k: ( B, \Theta_k ) \rightarrow B$, and $k(\cdot, \theta)$ is an endomorphism of the image function domain for each $\theta$. $k(\cdot; \theta) \circ s: \mathcal{D} \rightarrow B$ thus returns a template image with a unique spacing encoded in $\theta$. See the third item in Figure \ref{fig:completemodel}.

\subsection{Font and Appearance}

Now the canonical word image obtained by skeletonization has been spaced by the kerning function. At this point purely local transformations---{\it font} and {\it appearance}---produce the rendered text on a flat surface.

{\it Font} is the intrinsic shape of rendered text. For us, {\it appearance} consists of lighting variations, shadows, and imaging artifacts (such as blur). Appearance is an extrinsic feature while font is an intrinsic feature. We view the font as acting locally, widening curves, or elongating the skeleton characters at each point in the domain $I$. The $f$ function encodes local deformations mapping the skeleton to the font shape. 

{\it Appearance} comprises the effects of extrinsic features of the text location and environment. This includes background appearance and clutter, cast shadows, image noise, and rendering artifacts. However, appearance does not include geometric context or shape of the rendering surface in our model.

Font and appearance are independent of the word chosen, and we model the free parameters of font and appearance with a hidden variable domain $\Theta_f$. Font can be modeled by a function $f: (B(I), \Theta_f) \rightarrow B(I)$. To reflect the fact that disentangling appearance and font na\"ively is ill-posed, we model the appearance function $a$ on the parameter space $\Theta_f$.

\subsection{Geometric Distortion}

{\it Geometric distortion} of text arises primarily due to perspective distortion of planar geometry. The vast majority of text is rendered linearly onto planar surfaces, and so we adopt a homographic model for geometric distortion. In this work, we restrict ourselves to homogeneous linear models of geometric distortion to showcase the method. 

We fix the rendering domain to $R = [0,1, \ldots, 31] \times [0, 1, \ldots, 255]$ to enable batching for our training and to fix the dimensions of the input to our model. Note that the rendered image domain $I$ may or may not be aligned with $R$. The geometric distortion acts on $I$, so we may model it as parametrically pulling back $B(I)$ to $B(g(I))$: $ g(H) \ast (a\circ f \circ k \circ s) (w)|_x = [(a \circ f \circ k \circ s)(w)]|_{H(x)}$---where $H$ is a homography, and subscript indicates evaluation of the function at a point.

As mentioned above, STNs have been used in recent works toward modeling the geometry of the scene within an image~\cite{DBLP:journals/corr/ShiWLYB16,DBLP:conf/bmvc/LiuCWSH16}. However, modeling geometric distortion due to mismatches in the prediction space of the detector and the geometry of the scene are not handled in these previous works. Furthermore, we employ the new IC-STN~\cite{DBLP:conf/cvpr/ZhouYQJ17} which is capable of handling larger distortions by recursive computation of the distortion. 

\subsection{Complete Model}

The complete compositional model we propose models the text construction process by these 5 steps, given a word. The final image function is given by
\begin{align}
i = (g(H) \ast ((a\circ f)(\cdot, \theta_f) \circ k(\cdot, \theta_k) \circ s))(w).
\label{eqn:fullcompositionalmodel}
\end{align}

In Figure \ref{fig:completemodel} we show a module diagram that captures our compositional model for text. Next, we detail how the compositional model can be inverted and focus on modeling these functions. 

\section{Inverse Inference}

Inverting Equation \eqref{eqn:fullcompositionalmodel} is often done with an entangled combination of a CNN and an RNN. It is impossible to say theoretically which component models which aspect of the text and very difficult to quantify. We decompose 4 of these 5 processes into distinct functions and design an neural architecture to estimate these functions.

\subsection{Geometric Rectification}

Rectification can be implemented in a number of other ways, including  extracting and aligning keypoints~\cite{DBLP:journals/cviu/ChuiR03} or regions \cite{mikolajczyk2005comparison}, unsupervised aligning to a corpus~\cite{DBLP:conf/iccv/HuangJL07}, and image to image alignment \cite{DBLP:journals/ijcv/ViolaW97}. 

However, we have a particularly constrained scenario. It is difficult to establish a canonical pose for keypoints without formulating the skeletonization process to account for keypoint variation which introduces more free parameters and complexity. Image-to-image matching based on, for example, mutual information, is also difficult at this stage because the  appearance features have not yet been normalized. Finally, alignment to a mean image or a corpus of images involves painstaking congealing optimization which can make the solution slow and is difficult to train end-to-end~\cite{DBLP:conf/nips/HuangMLL12}. Thus, the region-based alignment methods and the STN are the most practical for this work. In this work we compare the affine patch-based MSER model \cite{mikolajczyk2005comparison} with the STN solution. Specifically, we estimate the MSER within a contextual region around each word detection, and then rectify to a minimal enclosing oriented bounding box for the MSER. We found that the STN provides a more reliable rectification module (See results in Figure \ref{fig:pertubationablation}).

We model geometric distortion by estimating a homography directly from the input pattern and coordinates of a given input region of an image. In the general case, there is no additional observation beyond the word pattern supervision, so we train the parameters to our rectifier in a weakly supervised manner outlined below. Thus, an inverse-composition STN (IC-STN) \cite{DBLP:conf/nips/JaderbergSZK15,DBLP:conf/cvpr/LinL17} is a natural fit. This form of geometric modeling allows for recapturing missing parts of the image unlike previous approaches~\cite{DBLP:journals/corr/ShiWLYB16,DBLP:conf/bmvc/LiuCWSH16}, and does not depend on fiducial point estimates~\cite{DBLP:journals/corr/ShiWLYB16}.
\begin{figure}[t!]
    \centering
    \begin{subfigure}[b]{0.25\textwidth}
        \centering
		\includegraphics[width=.5\linewidth]{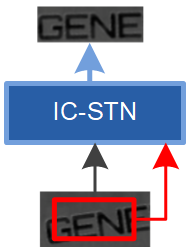}
        
        \caption{}
    \end{subfigure}%
    \begin{subfigure}[b]{0.25\textwidth}
        \centering
		\includegraphics[width=.7\linewidth]{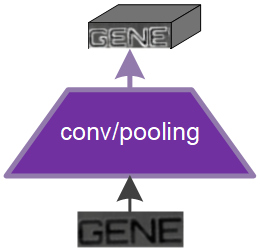}
        \caption{}
    \end{subfigure}%
    \caption{In subfigure (a) the IC-STN is pictured. Note that it takes both the coordinates and image as input. In subfigure (b) the convolutional encoding module is shown. The outputs of (a) are the inputs of (b).}
    \label{fig:icstnandconvpooling}
\end{figure}
    
As shown in Figure \ref{fig:icstnandconvpooling}(a), the input to our geometric rectification module is an input image an the coordinates of a text region. As output, it provides a rectified crop of the image. 

\subsection{Appearance featurization}

As mentioned above, disentangling $a \circ f$ without some additional supervision is ill-posed. We group the functions together and instead estimate $(a\circ f)^{-1}$. This can be done in a number of ways, but for most tasks  deep CNN features~\cite{DBLP:conf/nips/KrizhevskySH12,DBLP:conf/cvpr/HeZRS16} represent the state of the art in performance. Thus, we chose to model the font and appearance with Resnet-6. Figure \ref{fig:icstnandconvpooling} contains a visualization of the module.

\subsection{Kerning module}

While the receptive field of each of the kernels in our final feature layers contain an appreciable context, we believe that modeling kerning with the feature CNN leads to entangled modeling and thus reduced performance on this task with such a clear forward process. 

There are three main challenges to inverse inference of the kerning parameters: 
\begin{enumerate}
\item Spacing may require a variable amount of context within each image (due to distortion),
\item Spacing may require a variable amount of context for different images,
\item Features may not reflect the original spacing.
\end{enumerate}
We address Item 1 by the use of the geometric module in the front of the network. Thus, the features returned from the Resnet-6 module are in a canonical frame of reference. Items 2 and 3 we resolve by establishing an auxiliary loss that encourages locality of the features and by explicitly model the context around each spacing element using a bidirectional long-short-term memory (LSTM), respectively. We chose to use the LSTM because it is known to robustly model contextual relationships in ordered data.~\cite{hochreiter1997long} The auxiliary mean squared error loss is explained in more detail below. It helps to enforce consistency on the output of the kerning-LSTM (k-LSTM). We describe how the LSTM is implemented below.

\begin{figure}
\centering
\includegraphics[width=.8\linewidth]{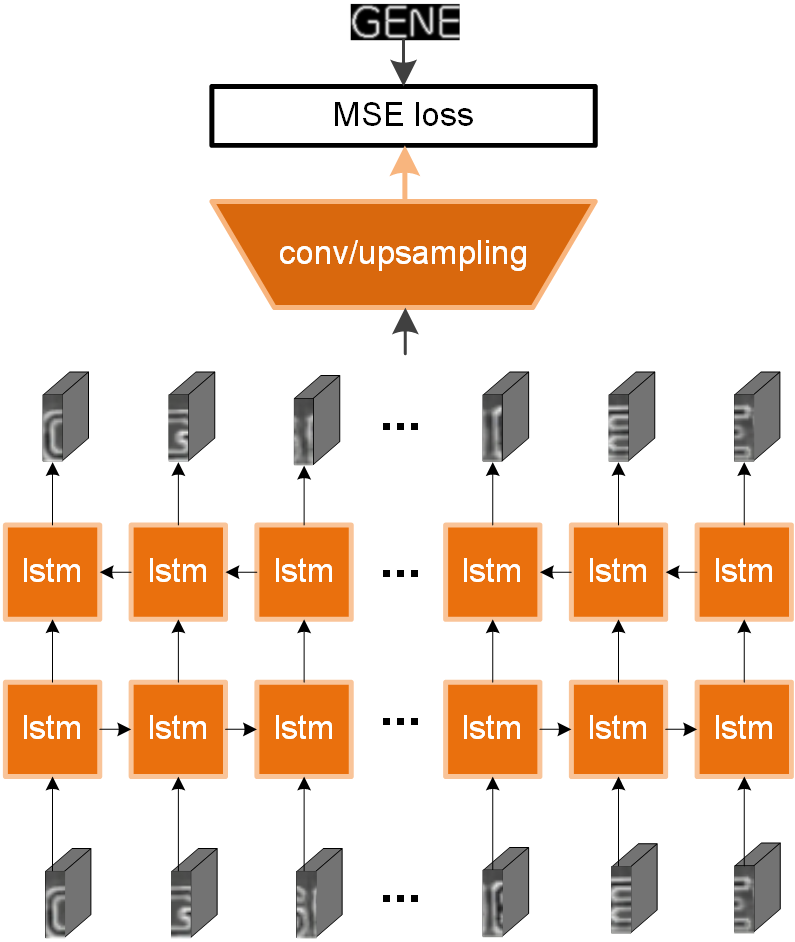}
\caption{The inputs to the kLSTM module come from the outputs of the convolutional encoder (see Figure \ref{fig:icstnandconvpooling} (b)). They are windowed as described in this section and the lstm is propagated forwards and backwards. The outputs of the bidirectional lstm are concatenated and fed into the convolutional decoder and the mean squared error loss is computed against the ground truth skeleton. }
\label{fig:klstm_mse}
\end{figure}

To employ an LSTM on the encoded text features, we begin by segmenting the features along the x-axis with a sliding window with width 2 and stride 1. We order the windowed features (collections of feature vectors in the sliding moving window) from left to right in increasing x order. This corresponds to the time index. The LSTM takes the vectorized features as input and computes a new internal state and then operates on the next window. After a pass over all of the feature windows, the bidirectional LSTM operates in reverse order and takes as input the output of a previous cell. Finally, the reverse cells output next feature vectors that are concatenated to form a new feature encoding. See the schematic in Figure \ref{fig:klstm_mse} for a visualization. This output is fed into the recognition and reconstruction modules.

\subsection{Skeleton reconstruction}

To rebuild the skeleton from the rectified and de-kerned features, we employ a deconvolutional architecture of 3 residual layers with spatial upsampling between each set of conv, conv, and add layers. This is designed to mirror the features in the encoding stage. Finally, a convolutional layer predicts the skeleton template. The ground truth template is compared with the prediction elementwise and the mean squared error is computed as outlined below.

\begin{figure}
\centering
\begin{tabular}{c c}
\includegraphics[width=.45\linewidth]{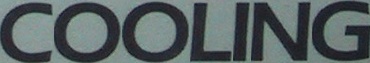} & \includegraphics[width=.45\linewidth, height=.08\linewidth]{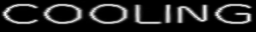} \\
\includegraphics[width=.45\linewidth]{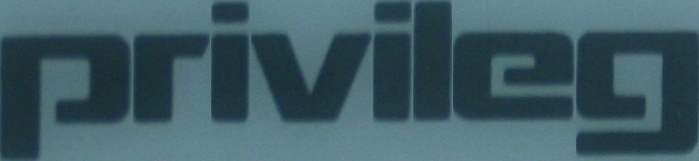} & \includegraphics[width=.45\linewidth, height=.08\linewidth]{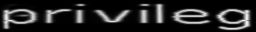} \\
\end{tabular}
\caption{Example images and templates.}
\label{fig:skeletonpics}
\end{figure}

We rendered templates for the given transcription of ground truth word images with the following process. Of course, a font must be chosen for the template; this will create a slight bias for the reconstruction task so we chose a standard font. First, we use the Sans font in GNU GIMP as template images. Then the skeleton $S_k$ of the character images is computed for each character $c_k$. Finally, the function $T_k:I \rightarrow R : x \mapsto \exp\{- d(x, S_k)^2/{2 \sigma^2 } \}$ is computed over a fixed image grid for each $k$. $\sigma$ is fixed to 1 pixel. Since there is some variation in the ligature between characters, as well as ascenders and descenders for each character, to fix kerning we resize all characters to consume the same amount of space in each template image (see Figure \ref{fig:skeletonpics}).

Using a template skeleton provides several advantages for inverse inference on all of the components above:
\begin{enumerate}
\item Registration can typically have trivial local minima without templates (normally MSE has trivial minima for joint registration, and a correlation-like loss must be employed), 
\item The kerning width is in terms of a fixed unit (the template kerning width),
\item Since the template skeletons are in terms of a fixed and clear font, legibility of the reconstruction implies that the features contain the information required to encode the characters for decoding. This provides another point of inspection for understanding the network.
\end{enumerate}

\subsection{Text recognition}

The text recognition module in our full model takes its input from the kerning module. We apply a residual layer and a convolutional layer to adapt to the recognition problem. Finally, an LSTM is used to convert the convolutional spatial features into character probabilities. Specifically, following~\cite{DBLP:journals/corr/ShiBY15} we use a bidirectional LSTM without peephole connections, clipping, or projection with stride 8 (1) in pixel space (feature space). The bidirectional LSTM enables contextual information to propagate along with the hidden state from a symmetric horizontal window along the columnar features. 

Below, we outline the functions used to drive the learning of the parameters of our inverse compositional modules.

\subsection{Loss functions}

We provide two objectives for our network: recognition and reconstruction. We review the CTC recognition loss~\cite{DBLP:conf/icml/GravesFGS06} first, then we overview and outline the reconstruction loss.

\subsubsection{CTC Recognition Loss} 

The CTC loss function operates on the output of the recognition LSTM. It produces a sequence of score vectors, or probability mass functions, across the codec labels. The conditional probability of the label given the prediction is the sum over all sequences leading to the label, given the mapping function $B$, 
\begin{align*}
p(l|y) = \sum_{\pi : B(\pi) = l} p(\pi|y). 
\end{align*}
This probability can be computed using dynamic programming~\cite{DBLP:conf/icml/GravesFGS06}.

CTC loss allows the inverted transcription process to have many-to-one correspondences. By marginalizing over the paths through the sequence prediction matrix that output the target, using dynamic programming, the  probability of the target sequence under the model parameters is obtained.

\subsubsection{Mean Squared-Error Reconstruction Loss} 
The MSE reconstruction loss takes as input the output of the decoder, the predicted template $\mathcal{S}:I\rightarrow R$, and the ground truth template image $\mathcal{T}:I\rightarrow R$ associated with the given word during training. See Figure \ref{fig:klstm_mse}. The objective function for this task is 
\begin{align}
L_{mse}(\mathcal{S}, \mathcal{T}) = \frac{1}{|I|}|| \mathcal{S} - \mathcal{T} ||^2_{L_2(I;R)} .
\end{align}

\section{Empirical Studies}

We study the performance of our recognition system both from the recognition standpoint and the end-to-end standpoint. Since we focus on the recognition step in this work, we have not emphasized a discussion of the detector design. We use a Faster-RCNN-based multiscale detector~\cite{DBLP:journals/corr/abs-1804-09003}. For comparison with existing end-to-end systems, the detector obtains recall of 90\% and F1-score of 92\% on the ICDAR Focused test set. In all experiments below, the detector model is frozen and is not used in any way to train the recognizer.  

\subsection{Ablation Experiments}
Our work features several new components, such as the template prediction loss and inverse compositional spatial transformer network. So we evaluate the performance of our model under several architectures and loss balancing parameters. This serves a validation for the significance of each component.

\subsubsection{End-to-end evaluation} 

In this experiment, we set the base learning rate to $1\times 10^{-4}$ used exponential decay of the learning rate with a factor of $0.9$ every 5000 iterations. The ADAM optimizer is used with $\beta_1 = 0.5$. We use the MLT Training and the ICDAR Focused Training datasets for training the recognizer. We perform data augmentation by the method described below, randomly perturbing the input coordinates, with a fixed $\sigma_p=0.025$.

\begin{table}[ht!]
\centering
\begin{tabular}{|c|c|c|c|c|}
\hline
Model & Generic & Weak & Strong \\
\hline
Baseline  & .798 & .847 & .861 \\
\hline
Full  & {\bf .824} & {\bf .862} & {\bf .872 } \\
\hline
Full-SSD  & .800 & .848 & .860 \\
\hline
Full-STN  & .804 & .856 & .869 \\ 
\hline
Full-STN+MSER & .641 & .765 & .763 \\ 
\hline
Full-kLSTM  & .800 & .854 & .868 \\
\hline
\end{tabular}
\caption{End-to-end F-scores for ICDAR 2013 Focused test set. This ablation experiment shows the significance of template reconstruction loss and spatial transformer and lstm-based kerning modules. All of these models are trained under identical settings.}
\label{table:ablation1}
\end{table}

\subsubsection{Robustness to Coordinate Perturbation}

Although data augmentation is often used for training text recognition algorithms, the robustness to perspective distortion is seldom studied in these works. Here we give a comprehensive evaluation of the robustness of our model under several ablations. 

\begin{figure}

\begin{tabular}{c c c}
\includegraphics[width=.3\linewidth]{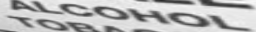} & \includegraphics[width=.3\linewidth]{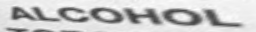} & 
\includegraphics[width=.3\linewidth]{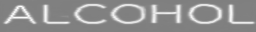} \\
\includegraphics[width=.3\linewidth]{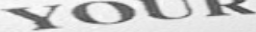} & \includegraphics[width=.3\linewidth]{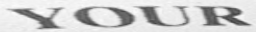} & 
\includegraphics[width=.3\linewidth]{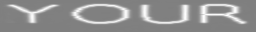} \\
\includegraphics[width=.3\linewidth]{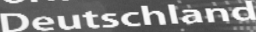} & \includegraphics[width=.3\linewidth]{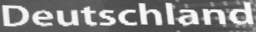} & 
\includegraphics[width=.3\linewidth]{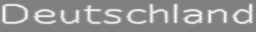} \\
\includegraphics[width=.3\linewidth]{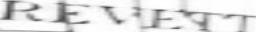} & \includegraphics[width=.3\linewidth]{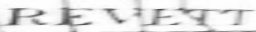} & 
\includegraphics[width=.3\linewidth]{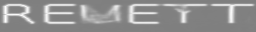} \\
\includegraphics[width=.3\linewidth]{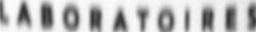} & \includegraphics[width=.3\linewidth]{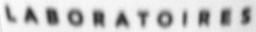} & 
\includegraphics[width=.3\linewidth]{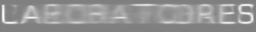} 
\end{tabular}
\caption{Example input images, rectified, and predicted templates. The first three images show examples that work well. The fourth example shows shortcomings of the featurization process, which indicates that explicitly modeling background may lead to improvements. In the last row, we see that out-of-model geometric transformations may also cause problems.}
\end{figure}
\begin{figure}[ht!]
\includegraphics[width=1.0\linewidth]{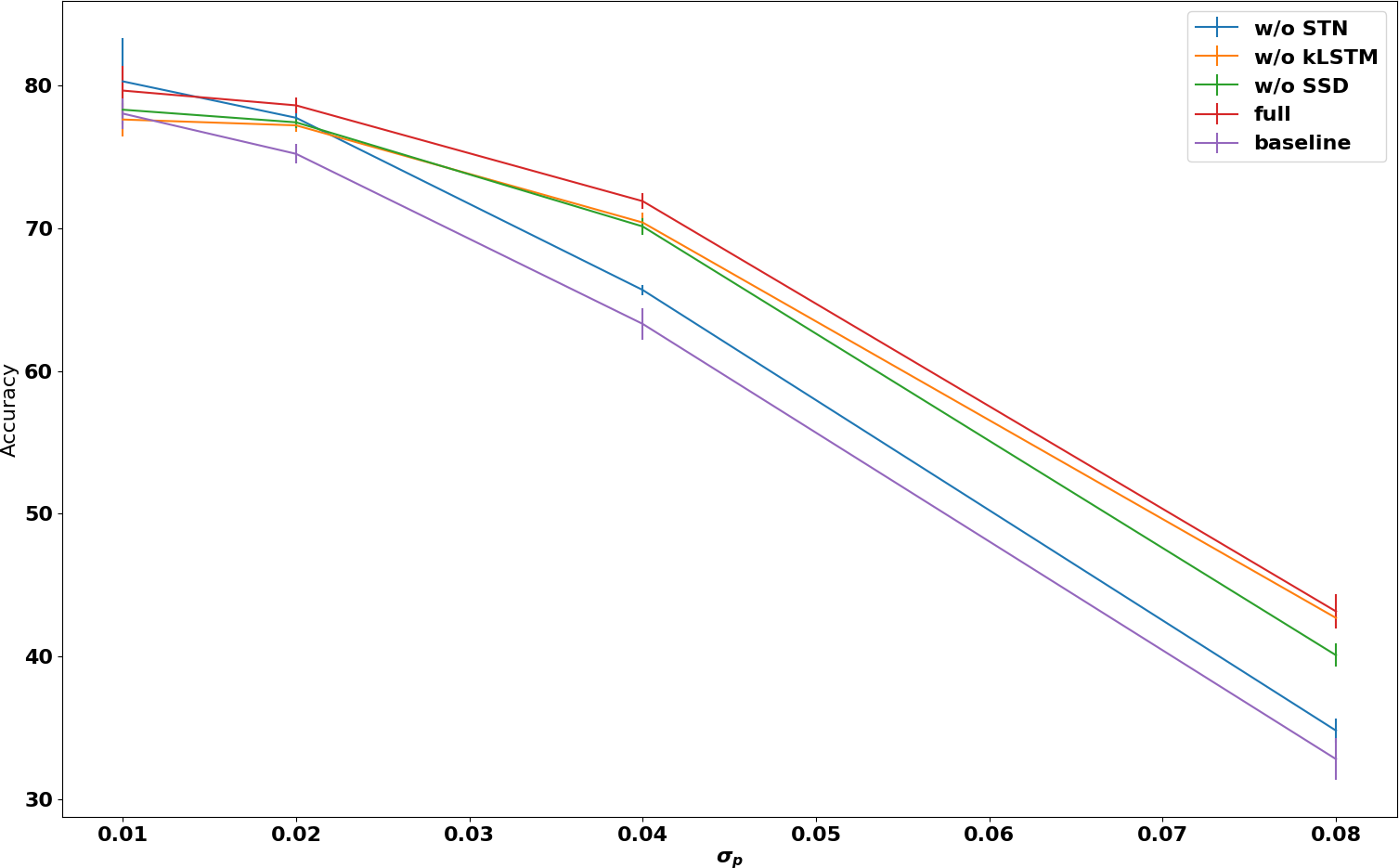}
\caption{The accuracy of the models from the ablation study under various degrees of perturbation of the ground truth coordinates. Notice the large gulf between models with and without an STN as the perturbation level rises. }
\label{fig:pertubationablation}
\end{figure}

For each input coordinate pair $X_i$ in each bounding box $B_j$, a vector $\eta_i \sim \mathcal{N}(0,\Sigma_p )$ is drawn. Then for each box $B_j$ a translation vector $t_j \sim \mathcal{N}(0,\Sigma_t )$ is drawn. Finally, the corrupted coordinates $\hat{X}_i^j = X_i + \eta_i + t_j$ replace the original coordinates in the input to our algorithm, for evaluation. $\Sigma_t, \Sigma_p$ are chosen to be a diagonal matrix $\Sigma_t = \sigma_t \Sigma, \Sigma_p = \sigma_p \Sigma$ with $\Sigma_{11} \propto w$ and $\Sigma_{22} \propto h$ with the constants of proportionality $\sigma_t=\sigma_p$ shown in the graph below, in Figure \ref{fig:pertubationablation}.

We performed 10 experiments at each noise setting and computed the mean and standard deviation of the accuracy for each model on each perturbed test set. In Figure \ref{fig:pertubationablation} error-bar plots are shown, in which the full model is clearly shown to improve significantly on the baseline, w/o STN, w/o SSD models for higher amounts of perturbation. This is further borne out in our comparative end-to-end experiments.

\subsubsection{Robustness to kerning variations}
We begin with the ICDAR-13 Text Segmentation dataset, which features character segments for each character in the word images. This provides us with an initial value for the spacing by computing the horizontal distance between the center of the enclosing bounding box of each segment. We then synthesize a sequence of images by reparameterizing the x-axis of the image so that the number of non-character columns between each character is increased by a factor of $1+0.1*k*H$ where $H$ is the height of the cropped word image.

In this section we provide qualitative anecdotal evidence that the k-LSTM module improves the quality of kerning modeling. 

\begin{figure}[ht!]
\includegraphics[width=\linewidth]{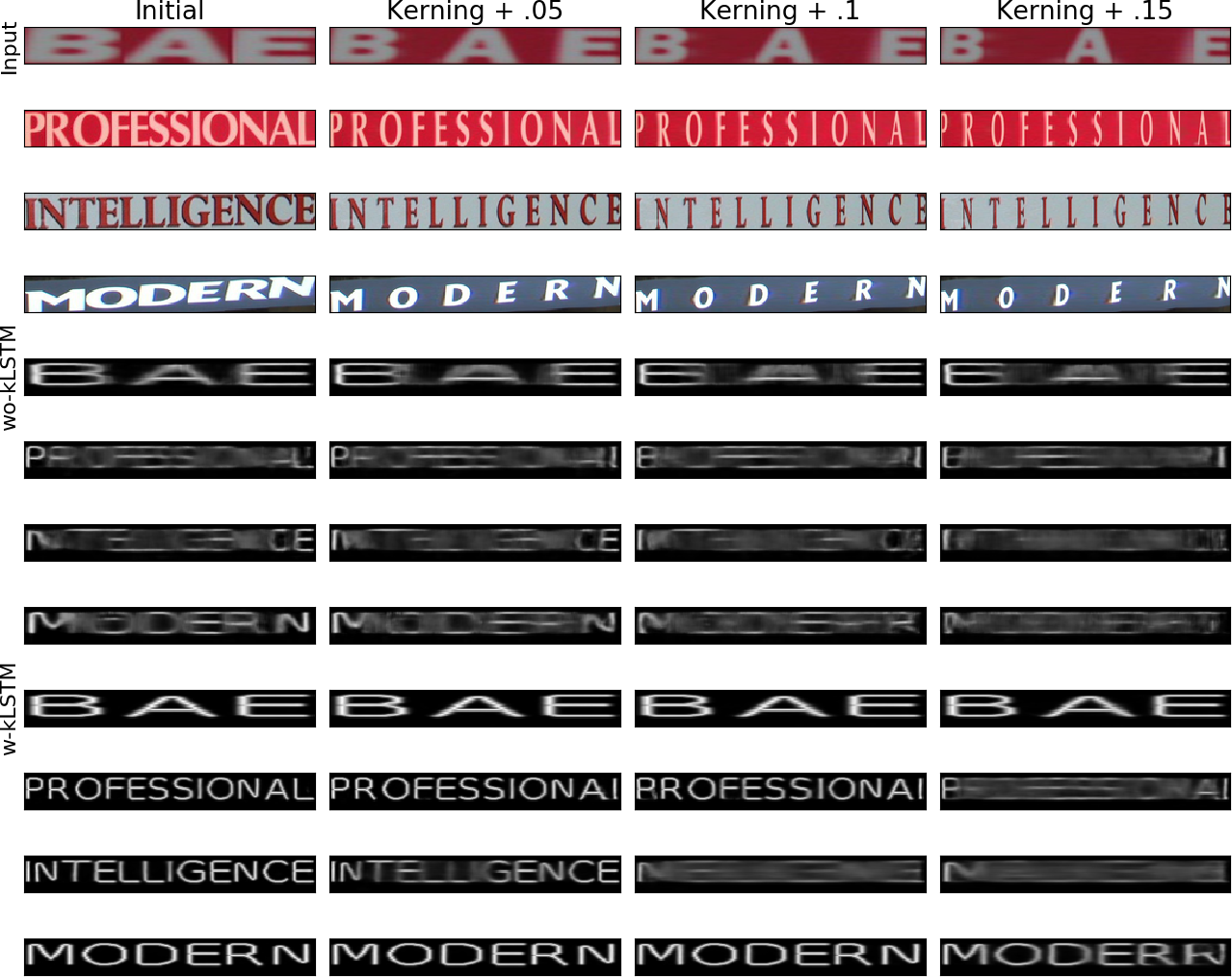}
\caption{In this image we show anecdotal evidence that a bidirectional LSTM can model the kerning process. In each column we show the input image (column 1) with increasing kerning. In the first four rows we show the input images. In the next four rows we show the predicted templates for a model w/o the LSTM. In the final four rows we show the predictions with the LSTM. While there are still failure cases for the full model, it improves significantly on the model without LSTM.}
\end{figure}

\subsection{Comparative Evaluation}

We compare our full model with state of the art text recognition algorithms and end-to-end algorithms on the ICDAR Focused 2013 and Incidental 2015 dataset. We used the same trained network for all experiments, with no additional fine-tuning, limited to the training data provided for the end-to-end task with the MLT 2017 training data in addition. For this experiment, we initialized our learning rate higher, to $5\times 10^{-4}$ since we observed that our full model is more robust to swings in the early training stages. Additionally, we employed a stagewise decay by a factor of $0.1$ after 10, 20, and 35  of the 50 epochs. We also incorporated a higher level of perturbation with $\sigma_p=\sigma_t=0.05$. 

\subsubsection{Recognition}

We test our network on Icdar Focused and Incidental for this task. We wanted to use SVT-P but we were unable to find a public hosting for this dataset. See Table \ref{tbl:recognitioncompare} for comparative results on the recognition tasks.

\begin{table}[t]
\begin{tabular}{|c|c|c|c|}
\hline
Algorithm & IC-13 & IC-IST \\
\hline
CRNN \cite{DBLP:journals/corr/ShiBY15} & 86.7 & - \\
\hline
RARE \cite{DBLP:journals/corr/ShiWLYB16} & 88.6 & - \\
\hline
Star-Net \cite{DBLP:conf/bmvc/LiuCWSH16} & 89.1 & - \\
\hline
Char-Net \cite{DBLP:conf/aaai/LiuCW18} & {\bf 90.8} & 60.0 \\
\hline
Ours & 89.4 & {\bf 65.0} \\
\hline
\end{tabular}
\caption{Accuracy with the 90K dictionary for the Icdar focused (IC-13) and incidental (IC-IST) recognition tasks. }
\label{tbl:recognitioncompare}
\end{table}

\begin{table*}
\centering
\begin{tabular}{|c|c|c|c|c|c|c|c|}
\hline
&
\multicolumn{3}{|c|}{ICDAR Focused} &
\multicolumn{3}{|c|}{ICDAR Incidental} 
\\
\hline
Algorithm & G & W & S & G & W & S  \\
\hline
TextBoxes++ \cite{DBLP:journals/tip/LiaoSB18}& 84.8 & {\bf 92.0} & {\bf 93.0}& 51.9 & 65.9 & 73.3 \\
\hline
RRPN* \cite{DBLP:journals/corr/MaSYWWZX17}  & 83.9 & 89.7 & 91.6 & - & - & 32.6  \\
\hline 
FOTS 2 Stage & 80.8 & 87.0 & 87.8 & 58.2 & 77.7 &80.4 \\
\hline
FOTS MS \cite{Liu_2018_CVPR} & 84.8 & 90.1 & 92.0 & {\bf 65.3} & {\bf 79.1} & {\bf 83.6}\\
\hline
Ours & {\bf 85.1} & 88.5 & 89.0 &57.7 & 72.4 &  75.6 \\
\hline
\end{tabular}
\caption{We show end-to-end performance for top OCR algorithms. The `G', `W', and `S' indicate the lexicon used for each task (Generic,  Weak, and Strong respectively), and the column shows the F-score of the respective algorithm. *We include the end-to-end number for RRPN from~\cite{Liu_2018_CVPR}.  }
\label{tbl:endtoendcompare}
\end{table*}

\subsubsection{End-to-end}

Even though our system is not trained end-to-end, we still have the best performance on the minimally assisted Generic end-to-end task. The modest improvements for the assisted tasks may be due to detector performance. However, since this work is showcasing the second component in the system we did not study this in depth. See Table \ref{tbl:endtoendcompare} for end-to-end comparative results.

\subsection{Conclusion}
OCR systems are known to be constrained by the first, detector, stage. Most previous works deal with this by coupling the recognizer and detector closer together and aligning the input of the second stage to the first. We offer an alternative strategy that involves training the recognizer to be capable of recovering missing text, and aligning the input patterns from the first stage for itself. We provided a full formulation for our approach based on a textual compositional model, and performed ablation to show the need for each component of the model. We provided comparative studies to place this work relative to the vast and quickly growing literature on text recognition and end-to-end OCR systems. 

We hope that in future work, the entangled modeling of font and appearance can be unraveled which may allow for further improvements. We also would like to explore parsimonious representations for nonlinear (homogeneously) geometric distortions in word images.

\newpage
\bibliographystyle{aaai}
\bibliography{aaai}
\end{document}